\documentclass{article}
\usepackage{ijcai19}

\usepackage{times}
\usepackage{amsmath}
\usepackage{amssymb}
\usepackage{graphicx}
\usepackage{array}
\usepackage{algorithm, algorithmic}
\usepackage{float}
\usepackage{stfloats}
\usepackage{soul}
\usepackage{url}
\usepackage[hidelinks]{hyperref}
\usepackage[utf8]{inputenc}
\usepackage[small]{caption}
\usepackage{graphicx}
\usepackage{amsmath}
\usepackage{booktabs}
\usepackage{algorithm}
\usepackage{algorithmic}
\usepackage{caption,subcaption}
\title{Deep Neural Architecture Search with Deep Graph Bayesian Optimization}

\author{
   \hspace{1cm}
}

\author{
Lizheng Ma$^{1,2}$
\and
Jiaxu Cui$^{1,2}$\and
Bo Yang$^{1,2}$\thanks{Corresponding author: ybo@jlu.edu.cn}
\affiliations
$^1${College of Computer Science and Technology, Jilin University, Changchun, China}\\
$^2${Key Laboratory of Symbolic Computation and Knowledge Engineering of Ministry of Education, China}
\emails
}

\begin{document}

\maketitle

\begin{abstract}
  Bayesian optimization (BO) is an effective method of finding the global optima of black-box functions. Recently BO has been applied to neural architecture search and shows better performance than pure evolutionary strategies. All these methods adopt Gaussian processes (GPs) as surrogate function, with the handcraft similarity metrics as input. In this work, we propose a Bayesian graph neural network as a new surrogate, which can automatically extract features from deep neural architectures, and use such learned features to fit and characterize black-box objectives and their uncertainty. Based on the new surrogate, we then develop a graph Bayesian optimization framework to address the challenging task of deep neural architecture search. Experiment results show our method significantly outperforms the comparative methods on benchmark tasks.

\end{abstract}

\section{Introduction}

\begin{figure*}[ht]
	\centering
	\includegraphics[scale=1]{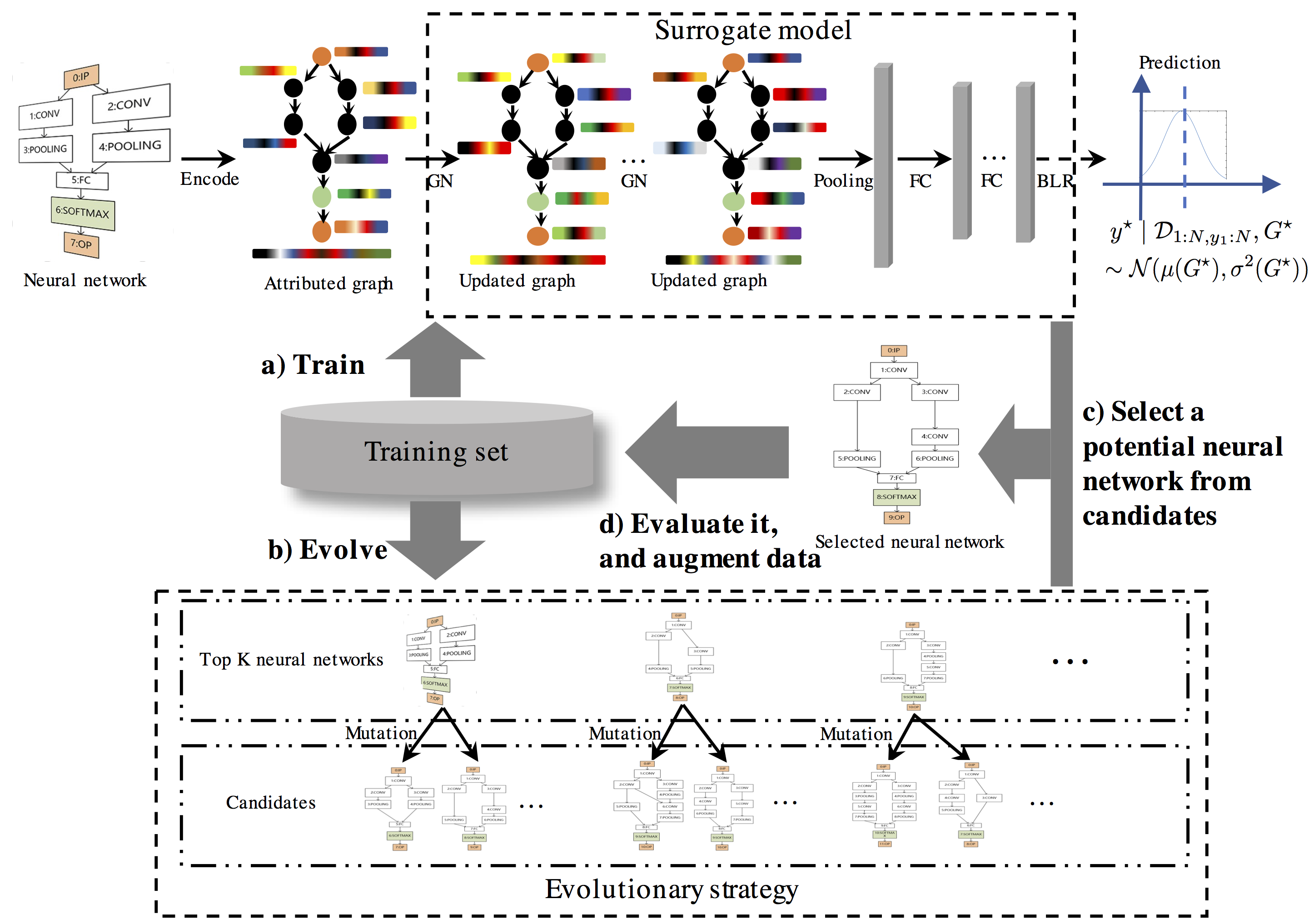}
	\caption{The workflow of the proposed NASGBO method.}
	\label{NASGBO}
\end{figure*}


Deep learning has been successfully applied in many fields including image recognition, speech recognition \cite{1hinton2012deep}, and machine translation  \cite{4sutskever2014sequence,5yonghui2016bridging}. To achieve good performance, these deep architectures need to be carefully designed by human experts. Due to the huge search space, it definitely takes great efforts to do this hard job. In recent years, there has been an increase of the literatures about neural architecture search \cite{19liu2018progressive,20liu2017hierarchical,12kandasamy2018neural}. Roughly, such works can be categorized into two main streams, one is based on reinforcement learning (RL) \cite{10zoph2016neural}, and the other is based on evolutionary algorithm (EA) \cite{20liu2017hierarchical,24real2018regularized}. EA-based methods can continually update the structures of neural nets to generate a series of generations for expanding search space through simpler and more efficient operations, and show promising results compared with other methods including RL-based ones. However, EA-based methods need to evaluate a large number of individual networks, which consumes a lot of expensive GPU resources.

To solve this problem, one can use Bayesian optimization strategy. The conventional solution of automatic machine learning resorts to formalizing machine learning process as a black-box optimization task. Bayesian optimization (BO) is an effective global optimization algorithm, with the goal of finding the optima of black-box objectives. Since the information about objective function is not known, BO utilizes a surrogate model to fit the black-box function, actively selects the most potential samples for real-world evaluations based on fitting results, and can quickly get the optimal location of the objective with only a few attempts. BO takes advantage of the information from the previous evaluations to update the quality of the surrogate model, which in turn, will help acquisition function to make better decision about where to evaluate next time. This is the main rationale why BO works more efficient. BO has a large number of applications including hyperparameter tuning for machine learning models \cite{25feurer2015efficient,26thornton2013auto}.


BO can be applied to improve the EA-based neural architecture search methods. By designing a good surrogate function, which can appropriately fit the classification or regression accuracy of a given neural architecture, BO can help EA-based method find an approximately optimal architecture by evaluating only a few candidates, and thus greatly reducing the computation overhead and saving GPU time. The current methods use Gaussian processes (GPs) as surrogate function, define the similarity between two neural architecture according to their structure difference, and then construct a similarity matrix as the kernel matrix of GPs \cite{12kandasamy2018neural,32jin2018efficient}.
Because the function mapping the structure of a neural architecture to its predictive performance is black box, it is difficult to know which features are appropriate or even relevant. The features that affect prediction may be structural features, attributes of nodes, global features of networks, or even some implicit features. Therefore, it is difficult for the man-made similarity matrix to reflect the characteristics of neural architecture in an all-round way, which maybe responsible for the failure of finding optimal solution. In addition, these manually customized kernel matrices contain subjective factors, and different definitions often lead to different results. Moreover, GPs is not scalable. Both the training and prediction of GPs involve computing the inverse of kernel matrix, with a time complexity of $O(N^3)$. When the number of observation samples $N$ becomes large, training and prediction become very slow, making it impossible to explore a larger search space.

In view of these problems, we plan to propose a new surrogate that can automatically extract useful features from neural architectures, and these features can be used to fit the mapping from architecture to predictive performance. Specifically, we model neural networks as attributed graphs and model the task of neural architecture search as the task of attributed graph optimization. Instead of GPs, we use Bayesian graph neural network (GNN) as new surrogate. GNN is a deep model of graph representation learning, which supports supervised learning of node and link embeddings from the context of attributed graphs. Its parameter sharing mechanism can greatly drop model complexity, which can not only reduce the time of training and prediction, but also avoid over-fitting. These advantages of GNN are especially suitable for the BO in which less training data are available. After embedding each layer, the representation of overall neural architecture can be obtained by pooling operations.

Based on the new surrogate, we develop a graph Bayesian optimization framework to address the problem of attributed graph optimization. We name it NASGBO, i.e., Neural Architecture Search with Graph Bayesian Optimization. Figure 1 illustrates its workflow. The proposed surrogate is given at the top. It consists of a GNN layer, a pooling layer, a MLP layer and a BLR layer. The input of GNN is the attributed graph encoding an input neural architecture, and its output is the embedding of nodes and links. Pooling layer combines node/link embeddings into the representation of the entire attributed graph and outputs it to MLP layer. MLP layer includes multiple fully connected layers, and predicts the predictive accuracy of the input neural architecture according to its embedding. In order to capture the uncertainty of MLP prediction, we add a BLR (Bayesian linear regression) layer. Note that we only add uncertainty at the last layer of the surrogate model, rather than modeling all model parameters as random variables. This is to balance the need of uncertainty measuring and the cost of computation.

By randomly generating, training and testing some neural architectures, we first prepare an initial training set. The training set is used to train the surrogate model (step a). Based on the current training set, a new population is generated by evolutionary operations (step b). The surrogate is used to predict the performance of each new individual, and one potential individual is selected out of them by maximizing the acquisition function (step c). Then the individual is trained, tested, and added to the current training set (step d). Under some cost constraints, the process is repeated until an approximate optimal solution is returned. Each component of the framework will be elaborated in remaining text.

%
%
%
%
%

\section{Problem Statement}

At present, there are two main ways to describe the architecture of neural networks \cite{28elsken2018neural}. One is multi-branch chained architecture \cite{12kandasamy2018neural} and the other is cell-based architecture   \cite{29liu2018progressive,30liu2018darts}. In the former representation, each layer can choose different operations such as pooling, convolution, etc. The output of each layer can be used as the input of all subsequent layers, not just as the input of the next layer. Resnets and Densenets belong to this type. This way is more flexible and can generate any kinds of network architectures, so the search space is correspondingly very large. In the latter representation, a neural network is constructed by cells as building blocks, e.g., normal cell and reduction cell. Because the structures of cells are fixed, one only needs to optimize the graphs of cells, so cell-based representation can greatly reduce the search space and allow us to generate deeper network architectures.
Limited by space, this work only focus on the multi-branch chained architecture as an example to demonstrate and verify our framework because it has larger search space with more challenges. It is worth noting that our proposed framework can be readily applied to the cell-based architecture by concentrating cells into nodes.

\subsection{Objective Formula}

Let $f \colon Q \to R$ be the performance evaluation function of neural architecture, where $Q$ denotes search space and $R$ denotes predictive accuracy. The goal of neural architecture search is to find the architecture with the highest predictive accuracy, so the objective function can be formalized as:
\[G ^\star = \mathop{\arg\max}_{q \in Q} \lbrack f(q) \rbrack\]
Each architecture in search space is modeled as an attributed graph $q=\lbrace V, E,F_{V},F_{G} \rbrace$, where $V$ is a set of nodes denoting the layers of neural architecture, $E$ is a set of edges, $F_{V}$ is the feature set of nodes and $F_{G}$ is the global feature set of the graph.

\subsection{Design of Search Space}

One of the key points is how to construct the neural network architecture search space $Q$. The search space defines the variables of the optimization problem and is also different corresponding to different search algorithms. In order to apply the GNNs method to Bayesian optimization, we use attributed graph to represent each neural network architecture. We encode the neural network into an attributed graph as input, which can fully exploit various features of the architecture. The attributed graph is mainly composed of network structure, layer attributes, and global attributes of the architecture. The network structure consists of a layer set $L$ and a directed edge $E$. The directed edge $(u,v)$  indicates whether the output of layer node $u$ is the input of the next layer $v$. Each node here is treated as a layer.

\begin{figure}[h]
	\centering
	\includegraphics[scale=0.9]{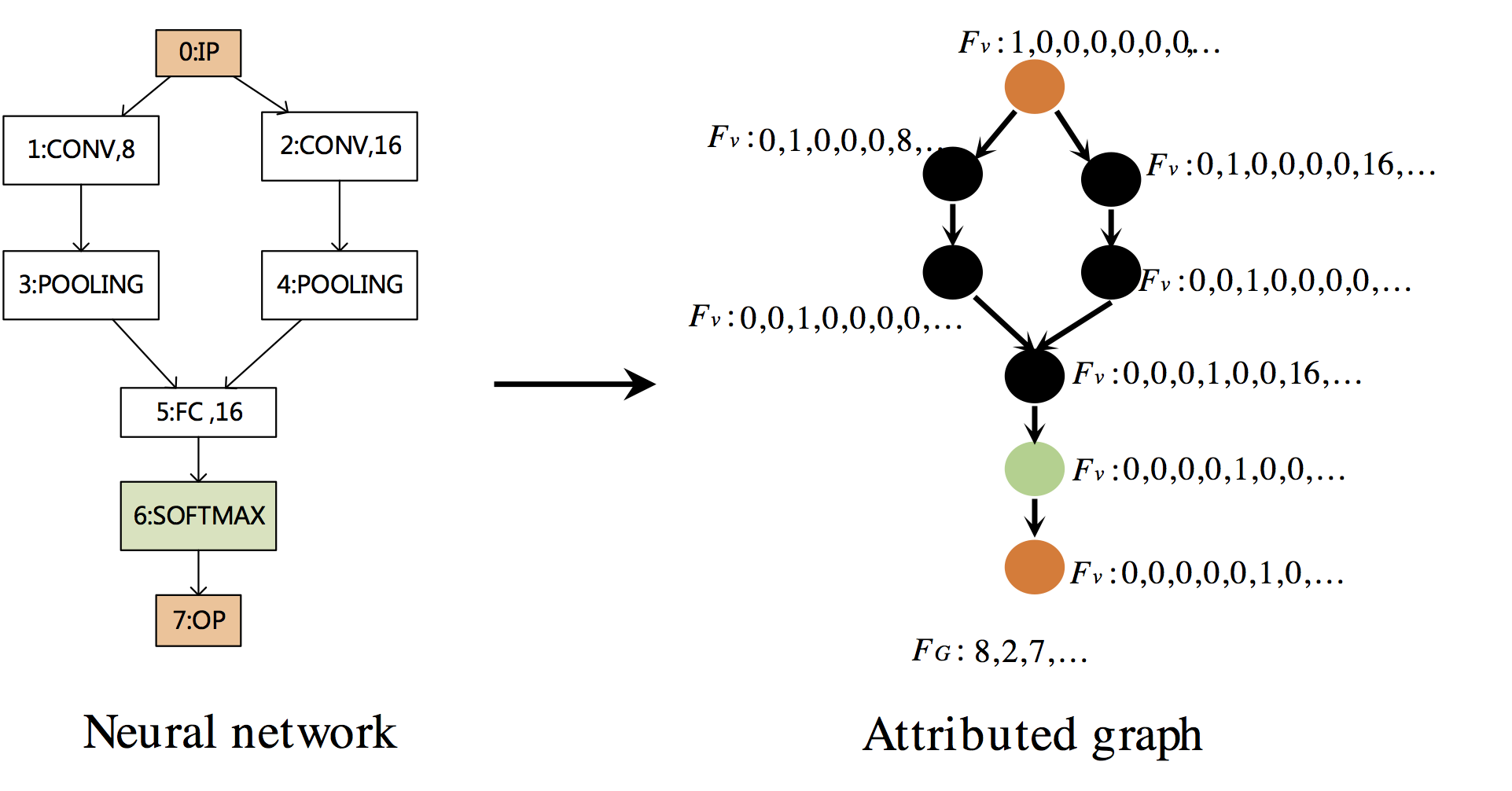}
	\caption{Representing a CNN as an attributed graph. $F_v$ and $F_G$  denote the node and global attributes, respectively.}
	\label{CNNs}
\end{figure}

Taking CNNs as an example, each layer is a node. Node attributes include layer types (softmax, conv3, conv5, conv7, res3, res5, res7, fc, max-pool, avg-pool, ip, op, etc), which each layer type occupys a single dimension in the node features. Other node attributes include the number of convolution units, the number of fully connected units, etc. The global graph feature include the num of the various node types, the num of nodes, the num of edges, etc. The connected edges between nodes represent the output of upper layer as the input of next layer. The index number of the node is obtained by an topological ordering, that is to say, a node with a smaller index number always points to a node with a larger index number.

Figure 2 provides an illustration from a CNN to the corresponding attributed graph. there are 8 nodes in the figure. To simplify the description, only first few dimensions of node attributes have been shown here, i.e. ip, op, cov3, pooling, fc, softmax, op and the number of units. For example, the attributes of node 2 include cov3 and the number 16 of units.

-point bold type. Leave a blank space

\section{The components of NASGBO framework}

As illustrated by the Figure 1, our proposed NASGBO framework consists of three main components: deep surrogate model,  acquisition function and evolutionary strategy. The workflow of NASGBO is described in Algorithm 1.

\begin{algorithm}
	\renewcommand{\algorithmicrequire}{\textbf{Input:}}
	\renewcommand{\algorithmicensure}{\textbf{Output:}}
	\caption{NASGBO}
	\label{alg:NASGBO}
	\begin{algorithmic}[1]
		\REQUIRE Neural network architecture initialization sets $G_0=\lbrace g_{1}, g_{2}, ... ,g_{m}\rbrace$, the architecture of GNNs surrogate model $SM$, hyper-parameter sampling $S$, iterations $Num$
		\ENSURE Optimal neural network architecture $g_{\max}$ and the performance $y$ of test set
		\STATE After the initialized neural networks $G_0$ are trained, get the performance $y_{i}$ of validation set, then integrate into $D=\lbrace (g_{1},y_{1}), (g_{2}, y_{2}), ... , (g_{m}, y_{m})\rbrace $, $G=\lbrace g_{1}, g_{2}, ... ,g_{m}\rbrace$.
		\FOR{$t = 1, 2, ... , Num$}
		\STATE Train $SM$ with training set $D$
		\STATE Sampling $S$ hyper-parameter samples from their posterior distribution $p(\theta \mid D)$
		\STATE $G$ generates $n$ individuals $G_{next}=\lbrace g_{1}, g_{2}, ... ,g_{n}\rbrace$ by evolutionary strategy
		\STATE Select a neural network architecture $g_{next}$ from $G_{next}$ by maximizing EI value
		\STATE After the $g_{next}$ is trained, get performance $y_{next}$ of validation set, then augment data $D=D \cup \lbrace (g_{next}, y_{next})\rbrace$, $G=G \cup \lbrace g_{next}\rbrace$.
		\ENDFOR
		\STATE Training the neural network architecture $g_{\max}$, where $g_{\max}$ has the best performance, get the performance $y$ of test set
		\STATE \textbf{return} $\lbrace g_{\max}, y \rbrace$
	\end{algorithmic}
\end{algorithm}



\subsection{Bayesian GNN surrogate}


Due to the issues of the GPs as mentioned in the introduction, we use a deep graph neural network as the surrogate function rather than GPs. In addition, to make our surrogate more scalable and be able to model uncertainty we integrate a layer of BLR (Bayesian linear regressor).

Graph Neural Networks (GNNs) is a deep learning method designed for graph data \cite{battaglia2018relational}, which can be used to learn the representation of an attributed graph.
GNN generalizes various neural network methods for manipulating graphs and defines a class of functions for relational reasoning based on graph structure representations. GN (graph network) module is the main computational unit of the GNN, which is a ``graph to graph'' module that takes the graph as the input and returns a graph as its output. The nodes and edges have their own attributes, which can be a vector or tensor.

In our framework, a GN block that we adopt contains three ``update'' functions:
\[e'_k = MLP_e([e_k,v_{r_k},v_{s_k},u]) \]
\[v'_i=MLP_v([\overline{e'_i},v_i,u])\]
\[u'=MLP_u([\overline{e'},\overline{v'},u])\]
where $\overline{e_i'}=sum(E_i^{'})$,
$\overline{e'}=sum(E^{'})$,
$\overline{v'}=sum(\overline{V'})$,
$MLP_e$, $MLP_v$ and $MLP_u$ have five layers,
 $sum$ is an element-wise sum operation, $u$ represents a vector of the global attributes, $V$ is a set of nodes, each $V_i$ represents the attributes of the node $i$, $E$ is a set of edges, where each $e_k$ represents the attribute of the edge $k$, $r_k$ is the index of the receiving node, and $s_k$ is the index of the sending node,
$E_i^{'}=\lbrace (e_k^{'}, r_k, s_k) \rbrace _{r_{k}=i,k=1:N^{e}}$, and  $V^{'}=\lbrace v_i^{'} \rbrace _{i=1:N^{v}}$. The edge attribute is updated with the node attribute connected to the edge, the node is updated with the edge attribute connected to the node.

Using the pooling layer, we make full use of the node features, the edge information of the graph and the global features to learn the global representation of the whole graph.In order to capture the uncertainty of the graph when predicting the metric of the graph, we add a Bayesian linear regression (BLR) as the last layer of the surrogate architecture behind multiple fully connected layers. We refer to this model as adaptive basis regression, which are parameterized by the weights and biases of the deep neural network. The form of the BLR is as follows:
\[y_{1:N}=\Phi(.)^{T}w+b\]

\noindent where y is the output of the surrogate function and $b \sim N(0,\sigma ^{2}_{noise}I)$,which is normal distribution, and $\Phi(.)$ is the decision matrix output by previous layers as the input of BLR layer. Given a prior distribution on weights: $w \sim N(0,\sigma_w^{2}I)$, where $\sigma_w^{2}$ denotes the uncertainty of $w$.

The measure of $G^{\star}$ can be predicted by:
\[y^{\star} \mid \mathcal{D}_{1:N,y_{1}:N}, G^{\star} \sim \mathcal{N}(\mu(G^{\star}),\sigma ^{2}(G^{\star}))\]
where $\mathcal{D}_{1:N}$ are observations, $y_{1}:N$ are evaluated measures,
\[ \mu(G^{\star})=\sigma^{-2}_{noise}\Phi(G^{\star})^{T}K^{-1}\Phi(.)_{y_{1:N}} ,\]
\[ \sigma^{2}(G^{\star})=\Phi(G^{\star})^{T}K^{-1}\Phi(G^{\star})+\sigma^{2}_{noise} ,\]
\[ K=\sigma^{-2}_{noise}\Phi(.)\Phi(.)^{T}+\sigma^{-2}_{w}I .\]

For Bayesian optimization, if surrogate function is the GPs, Maximizing the acquisition function takes $O(N^{3})$ time to calculate the inverse of surrogate matrix $(N\times N)$.For Deep Graph Bayesian Optimization, it takes $O(M^{2}N)$ and $O(M^{2})$ time to maximize the acquisition function. M is the number of units on BLR layer(usually $M\ll N$). Therefore, the Deep Graph Bayesian Optimization takes less time when maximizing the acquisition function.

\subsection{Acquisition function}
 In Bayesian optimization, it can be seen that the probability description of objective function $f$ can be quantified by sampling. Usually the acquisition function mainly uses exploiting and exploring sampling ideas. Exploring new spaces helps to estimate a more accurate objective function $f$, and sampling near the existing results (usually near the maximum) are expected to find a larger one. The purpose of the acquisition function is to balance the two sampling ideas and quantify the potential of candidate graphs based on previous validations.
Given a graph search
space $\mathcal{G}$ and a hyper-parameter space ${\Theta}$, we can define the acquisition function $\mathcal{U} : \mathcal{G}\times \Theta \to \mathbb{R}$. Although any other acquisition functions can be used in the proposed framework, this paper uses expected improvement (EI), a simple, valid, and common criterion, as the acquisition function. The EI function is the expectation of the improvement function $\emph{I}(\emph{\textbf{x}}^*)=max\{0,(\mu(\emph{\textbf{x}}^*)-y_{max})\}$ at candidate point $\emph{\textbf{x}}^*$ . Specifically, it can be fulfilled by
\[\mathcal{U}(\emph{\textbf{x}}^*|\mathcal{D}_t,{{\theta}})=(\mu(\emph{\textbf{x}}^*)-y_{max})\Phi(z(\emph{\textbf{x}}^*))+\sigma(\emph{\textbf{x}}^*)\phi(z(\emph{\textbf{x}}^*)),\]

\noindent where $ z(\emph{\textbf{x}}^*)=\frac{\mu(\emph{\textbf{x}}^*)-y_{max}}{\sigma(\emph{\textbf{x}}^*)}$, $y_{max}$ is the maximum value in the current set of observations $\mathcal{D}_t$, and $\Phi(.)$ and $\phi(.)$ denote the cumulative distribution function and probability density function of the standard normal distribution, respectively.

\subsection{Evolutionary strategy}

Unlike NASBOT and NASNM, We firstly use an evolutionary algorithm (EA) to generate neural networks of $k$ generations rather than optimising the acquisition function .
The value of $k$ may be 1, 2, etc. Then we use deep graph bayesian optimization to select the most potential neural network . However , by using EA to optimise the acquisition function , the selected neural network may be very deep and lead to resource crunch after many generations of mutation.
This is the main reason why we use EA only for generating. The above is just a simple strategy to balance the breadth and depth of mutation. It can also be tackled in other good strategies.

Each neural network is a feasible solution in the search space, and an approximate evaluation is needed to judge the performance of neural network.
We encode each neural network model into a graph. The mutation operation of a neural network model can be transformed into operation on the graph,
such as adding a node, deleting a node, adding an edge, etc. For example, adding a convolution layer is equivalent to adding a node to the graph.
The training environment and mutation operations of NASGBO are the same as NASBOT.
Table 1 shows several types of operations.

\renewcommand\arraystretch{1.3}
\begin{table}[h]
	\centering
	\begin{tabular}{cm{5cm} cp{5cm}}
		\hline
		Mutation Operation & Description\\
		\hline
		Adding skip & Randomly select two convolution layer ids for skip connection\\
		Incresing units &Increase the number of units by 1/8 \\
		Decresing units &  Decrease the number of units by 1/8\\	
		Adding layer & Randomly select two convolution layer ids for adding a convolution layer\\	
		Removing layer & Randomly select a convolution layer id for removing\\
		\hline
	\end{tabular}
	\caption{Descriptions of neural architecture mutation operations}
\end{table}

\section{Experiments and Analysis}

To evaluate the performance of the proposed Neural Architecture Search with Graph Bayesian Optimization (NASGBO), we compare it with the following baselines. We mainly verify the performance of NASGBO from two perspectives including accuracy and cost.

\subsection{Baseline Algorithms}

\begin{itemize}
	\item \textbf{RAND}: In the case of the same initial individuals, rand algorithm selects one from initial individuals for mutation and evaluation.
	\item \textbf{TreeBO \cite{15jenatton2017bayesian}}: A BO method that only searches over feedforward structures.
	\item \textbf{SEAS \cite{31elsken2017simple}}: A method to efficient architecture search for convolutional neural networks  based on hill climbing.
	\item \textbf{NASNM \cite{32jin2018efficient}}: A Bayesian optimization algorithm in which the network morphism is used to construct kernel function.
	\item   \textbf{NASBOT \cite{12kandasamy2018neural}}:A Bayesian optimization algorithm for neural architecture search. Firstly, NASBOT utilizes Optimal Transport Metrics for Architectures of Neural Networks to represent the similarity of the networks, and then searches neural network architectures through the GPs.
\end{itemize}

Unlike NASBOT and NASNM, we use graph neural network (GNNs) as the surrogate function. It can adequately represent the architectural features of the neural networks.

\subsection{Dataset}
We use five data sets for the experiments, as follows: Indoor Location \cite{13torres2014ujiindoorloc}, Slice Localisation \cite{11graf20112d}, Cifar10 \cite{21krizhevsky2009learning}, Minist \cite{2lecun1998gradient}, Fanshion Minist \cite{27xiao2017fashion}. The first two data sets are applied to the regression problem of MLPs. The last three data sets are applied to the classification tasks of CNN images.
For the first two data sets, we split the data sets using a scale of 0.6-0.2-0.2, which used as training data sets, test data sets, and validation data sets respectively, and normalize these data sets to have zero mean and unit variance.
For the Cifar10 data sets, there are 60,000 images, of which 50,000 are for training and 10,000 are for testing. We used a 40K-10K-10K ratio to segment the data sets, which used as training dataset, test dataset and validation dataset.
For the last two data sets, there are 70,000 images, of which 60,000 are for training and 10,000 are for testing. We used a 50K-10K-10K ratio to segment the data sets.

\subsection{Results}

Our method is executed on a single GPU (NVIDIA GeForce GTX 1080 Ti). The regression MSE or classification error (lower is better) on the test set is selected as the evaluation metric.

\begin{figure}[h]
	\centering
	\includegraphics[scale=0.41]{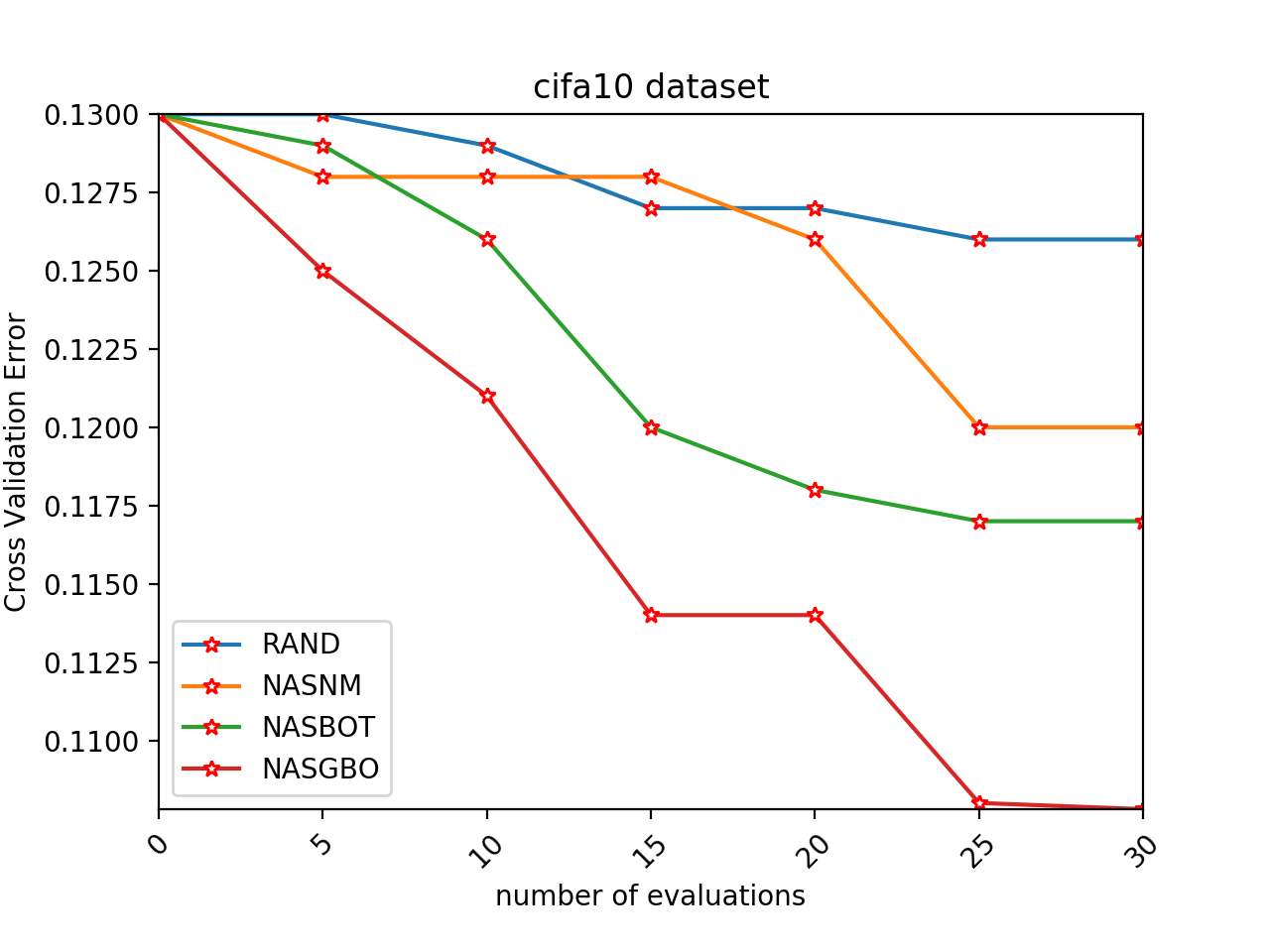}
	\includegraphics[scale=0.41]{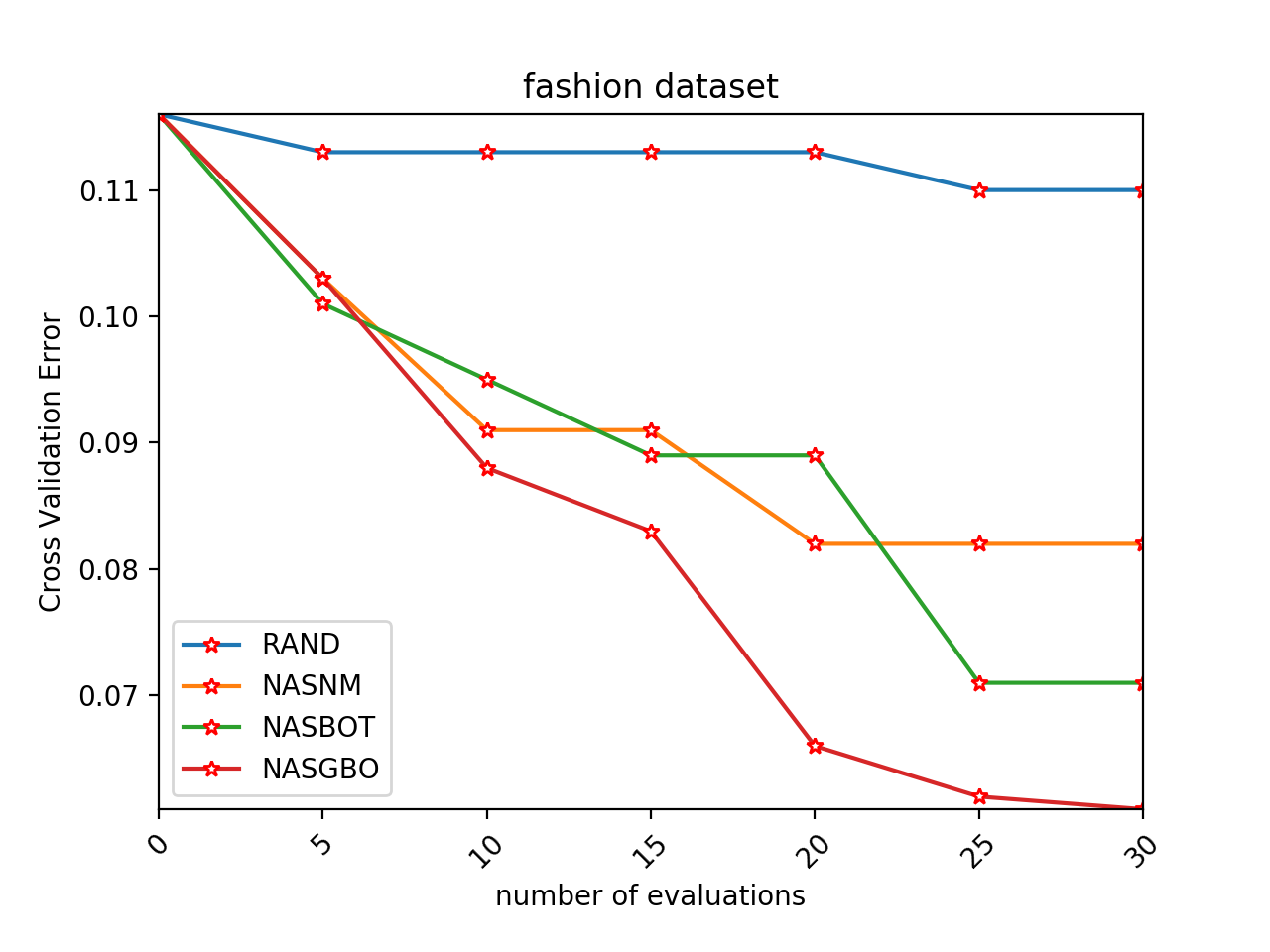}
	\includegraphics[scale=0.41]{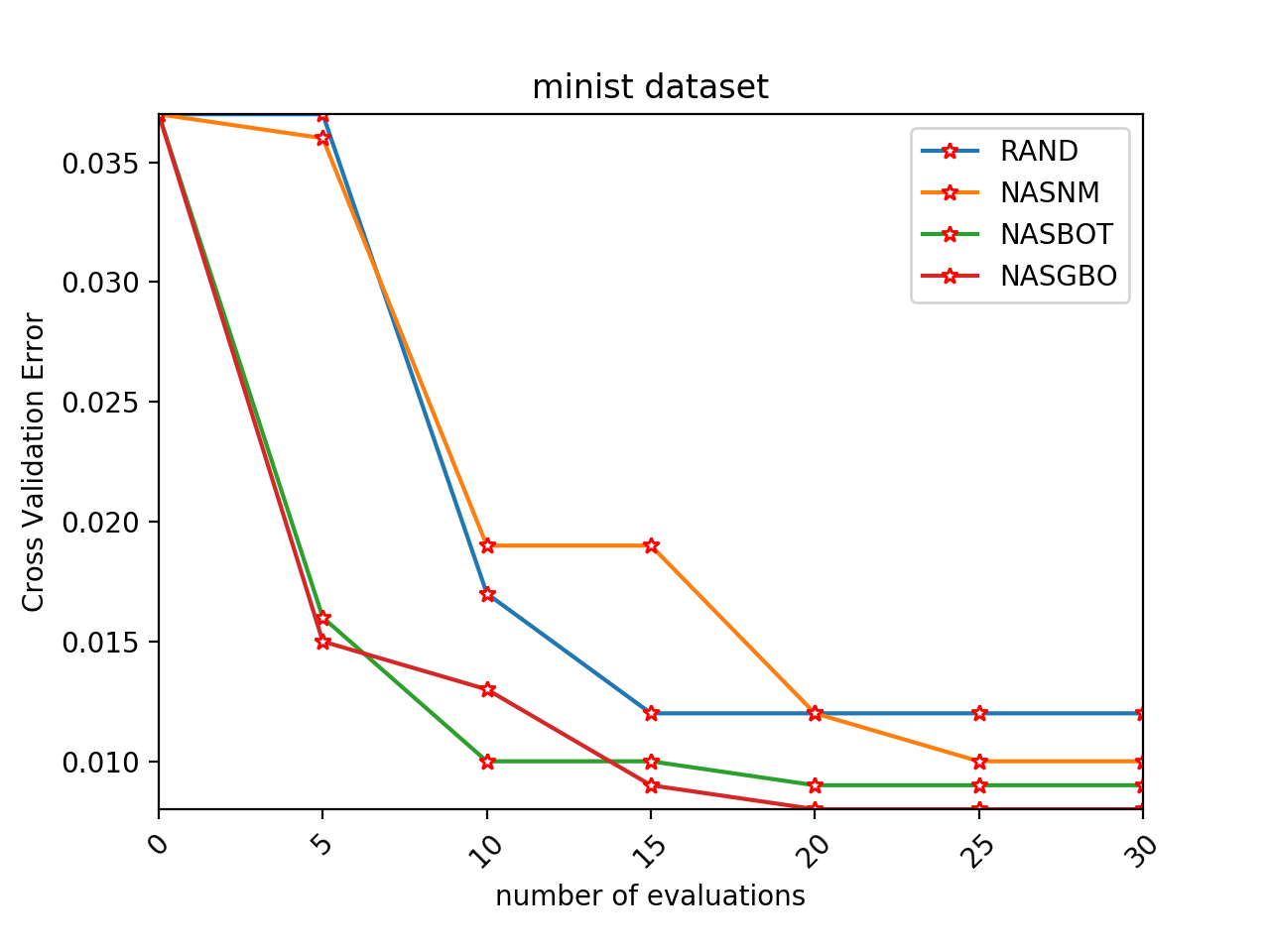}
	\caption{The best validation score for each method against the number of evaluations for CNNs}
	\label{The Components of NASGBO Framework}
\end{figure}

\newcommand{\tabincell}[2]{\begin{tabular}{@{}#1@{}}#2\end{tabular}}
\renewcommand\arraystretch{1.5}
\begin{table*}
	\centering
	\begin{tabular}{p{1.5cm}<{\centering}p{2cm}<{\centering}p{2cm}<{\centering}p{2cm}<{\centering}p{2cm}<{\centering}p{2cm}<{\centering}}
		\hline
		Method&Cifar10&Fashion&Minist&Indoor&Slice\\
		\hline
		RAND  &0.145&0.1100&0.012&0.156&0.932\\
		
		TreeBO&0.153&-- &-- &0.168&0.759\\
		
		SEAS&0.197&0.0800&0.013&--&--\\
		
		NASNM&0.123&0.0757&0.010&0.112&0.870\\
		
		NASBOT&0.122&0.0761&0.009&0.114&0.615\\
		\hline
		NASGBO&\textbf{0.120}&\textbf{0.0670}&\textbf{0.008}&\textbf{0.090}&\textbf{0.560}\\
		\hline
	\end{tabular}
	\caption{The subsequent rows show the regression MSE or classification error (lower is better) on the test set for each method.
	For TreeBO and SEAS, we choose the best results of test set from  NASNM.  "--" indicates that the program did not provide an experiment. }
\end{table*}

\begin{figure}[h]
	\centering
	\includegraphics[scale=0.42]{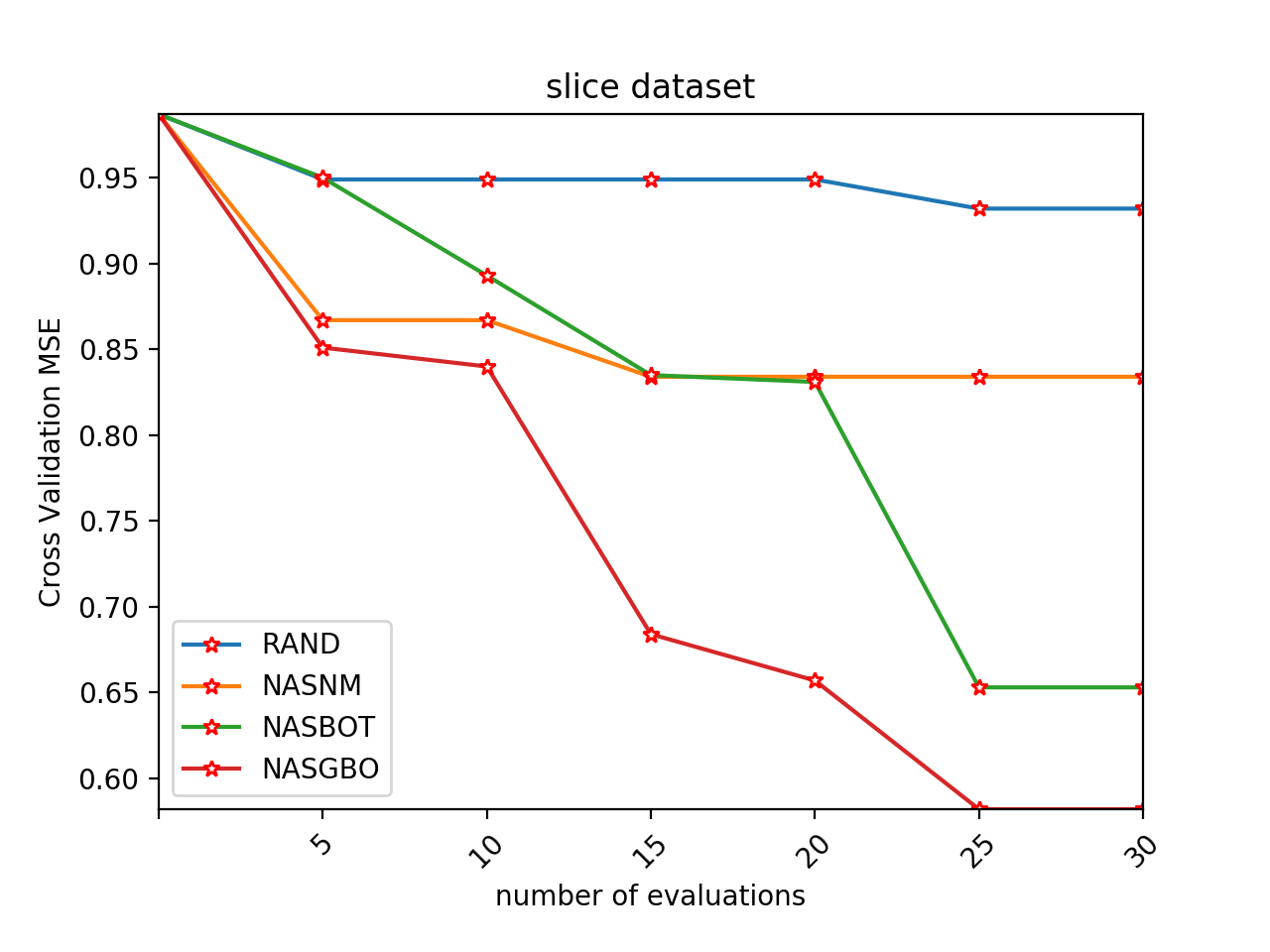}
	\includegraphics[scale=0.42]{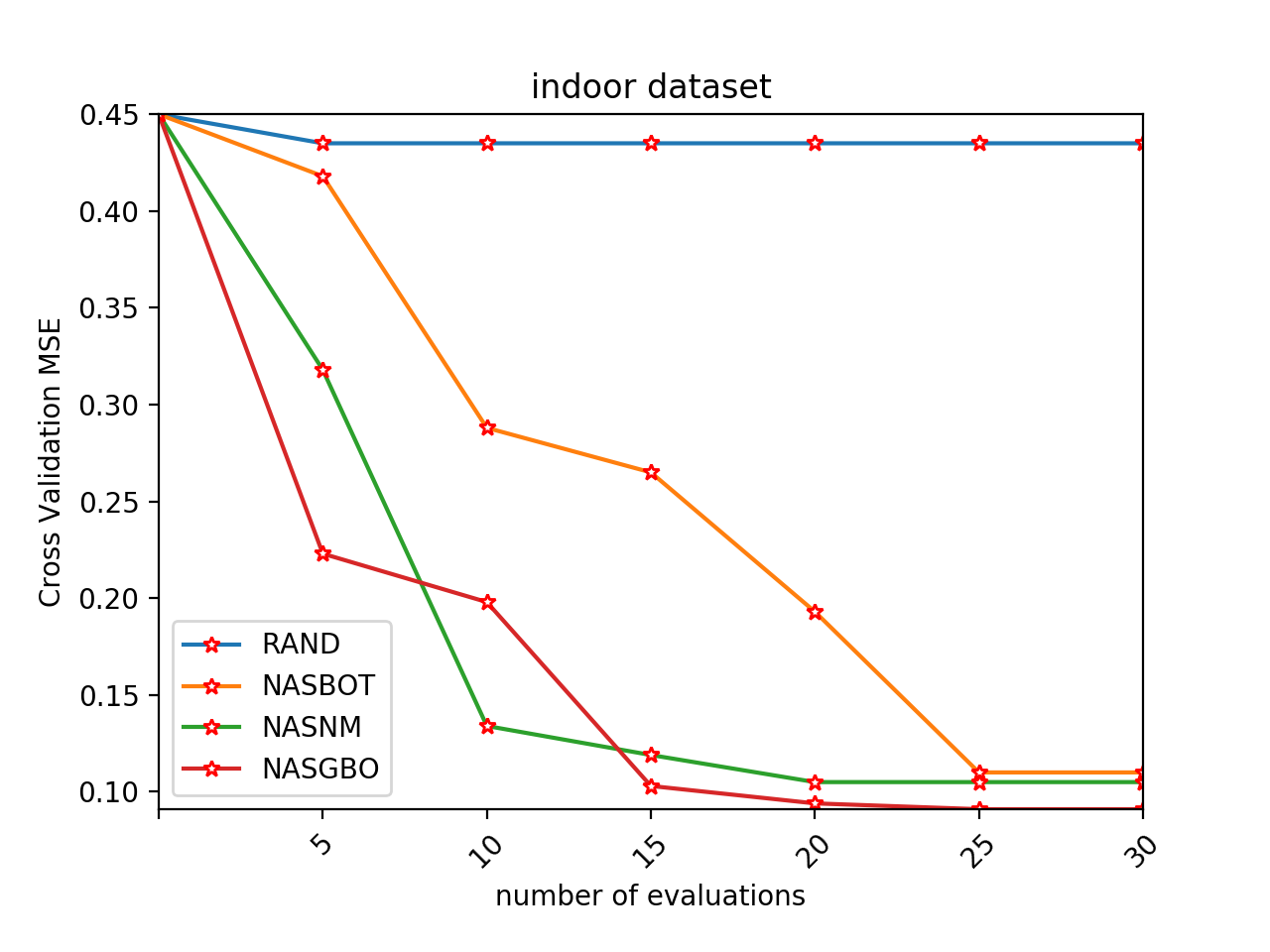}
	\caption{The best validation score for each method against the number of evaluations for MLPs}
	\label{The Components of NASGBO Framework}
\end{figure}

After running for a period of time (such as 12 hours), the corresponding accuracy of neural network may not increase in the next period of time. Finally, we train the best neural network on the test set to get results of test set.

Table 2 shows that the results on the test set with the best model. As shown by Table 2, NASBGO can get the best result of test set on Cifar10, Fashion, Mnist, and Slice data sets. NASBOT can get the best result on Mnist and Indoor data sets. Relatively speaking,  NASBGO performs better than other algorithms in getting the best network architecture.

We use the number of evaluations as cost metric, which does not contain the number of initialized neural networks. The training environment and mutation operations of NASGBO is consistent with NASBOT. We implemented NASBOT conscientiously, and use it as comparative experiment against the number of evaluations on the five data sets .Rand and EA algorithm can be implemented easily compared to our algorithm. We also add them to the comparative experiment.

Figure 3 shows best validation score for each method against number of evaluations for CNNs. As shown by the Figure 3, NASGBO converges faster than NASBOT on Cifar10 and Fashion data sets in terms of the number of evaluations. The performance of two algorithms is almost the same on Minist data sets.

Figure 4 shows best validation score for each method against number of evaluations for MLPs. As shown by the Figure 4, NASGBO converges faster than NASBOT on slice. For the Indoor data set, NASBOT is slightly better than the algorithm we proposed.

For CNNs and MLPs, we analyze the results of optimal architectures we obtained through fashion dataset and slice dataset. Figure 5 shows optimal architectures  of  NASNM, NASBOT and NASGBO on fashion dataset and Figure 6 shows optimal architectures  of  NASNM, NASBOT and NASGBO on slice dataset. 

As shown by the Figure 5, NASNM uses 5 residual blocks, NASBOT does not use any residual block, and NASGBO uses two residual blocks. From the overall perspective, we get a simpler architecture. According to table 2, we can see that classification error on the test set  is lower. 

For the slice dataset, the initialized architectures are straight-chain without skip connections. From the results, main reasons for affecting the results of slice dataset are the size of architecture and the skip connection of architecture. As shown by the Figure 6,  NASNM can learn about the impact of skip connections on the architecture, and add many skip connections. However, as the size of the architecture grows, the results get worse. NASBOT can also learn a simpler architecture than NASNM, but it is still relatively complicated compared to the architecture we have learned and the value of RMSE is higher than ours. With the use of skip connections, our algorithm makes the architecture as simple as possible. Finally, we can get better result on slice dataset from table 2.

Overall, our method can achieve competitive accuracy with less overhead for on MLPs and CNNs, the two benchmark neural architecture search tasks.
\begin{figure}[h]
	\centering
	
\begin{minipage}{0.48\linewidth}
\centerline{\includegraphics[scale=0.5]{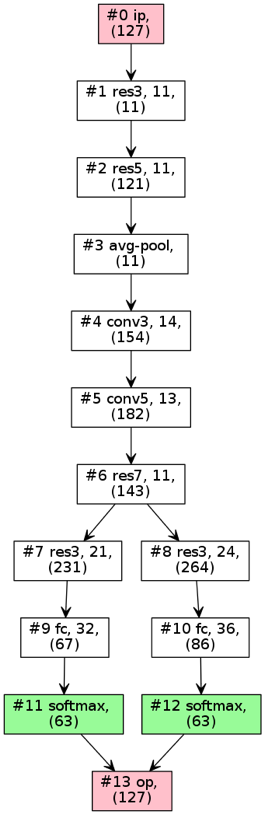}}
\centerline{NASNM}
\end{minipage}
\begin{minipage}{0.48\linewidth}
\centerline{\includegraphics[scale=0.5]{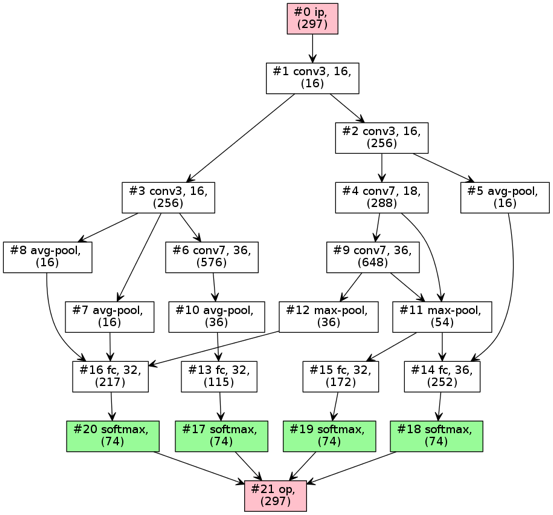}}
\centerline{NASBOT}
\end{minipage}
\begin{minipage}{0.48\linewidth}
\centerline{\includegraphics[scale=0.5]{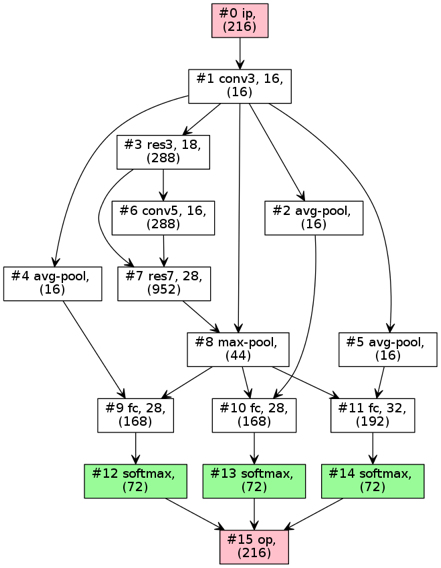}}
\centerline{NASGBO}
\end{minipage}

	\caption{Optimal architectures  of  NASNM, NASBOT and NASGBO on fashion dataset  }
\end{figure}

\begin{figure}[ht]
	\centering

\begin{minipage}{0.48\linewidth}
\centerline{\includegraphics[scale=0.5]{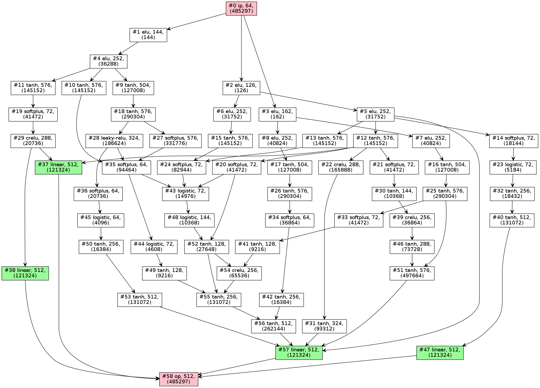}}
\centerline{NASNM}
\end{minipage}
\begin{minipage}{0.48\linewidth}
\centerline{\includegraphics[scale=0.5]{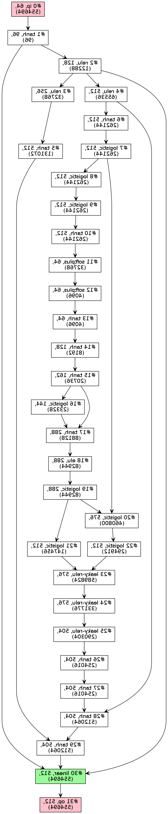}}
\centerline{NASBOT}
\end{minipage}
\begin{minipage}{0.48\linewidth}
\centerline{\includegraphics[scale=0.5]{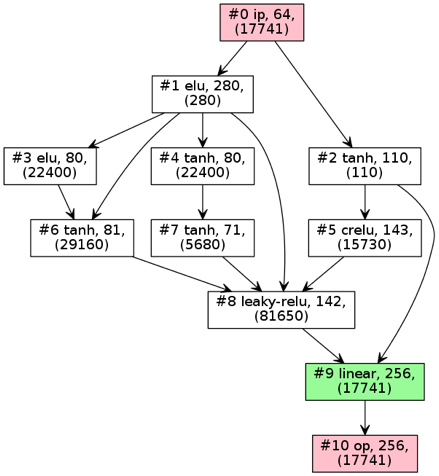}}
\centerline{NASGBO}
\end{minipage}

	\caption{Optimal architectures  of  NASNM, NASBOT and NASGBO on slice dataset  }
\end{figure}

\section{Conclusion}
The main contributions of this work are two-fold.
(1) We model neural networks as attributed graphs and model the task of neural architecture search as the task of attributed graph optimization. Based on this idea, we develop a graph Bayesian optimization framework to address neural architecture search, which integrating the advantages of EA-based methods and BO-based methods by designing a new surrogate model based on Bayesian GNN.  This new surrogate can automatically extract features from deep neural architectures, and can use such features to fit and characterize the black-box mapping from architecture to prediction as well as its uncertainty. Moreover, the training and prediction time of Bayesian optimization can be reduced to linear time by using the  Bayesian GNN surrogate, instead of the cubic time of Gauss process, which make our method be able to explore much larger search space and find better solutions.
(2) We rigorously show that our method outperforms the state-of-the-art works on benchmark neural architecture search tasks.

\newpage
\bibliographystyle{named}
\bibliography{ijcai19}

\end{document}